\documentclass[conference,10pt]{IEEEtran}
\IEEEoverridecommandlockouts
\usepackage{amsmath,amssymb,amsfonts}
\usepackage{algorithm}
\usepackage{algpseudocode}
\usepackage{array}
\usepackage{textcomp}
\usepackage{stfloats}
\usepackage{url}
\usepackage{color}
\usepackage{verbatim}
\usepackage{graphicx}
\usepackage{tabularx}
\usepackage{multirow}
\usepackage{array}
\usepackage{booktabs}
\usepackage{graphicx}
\usepackage{parskip}

\usepackage[numbers]{natbib}
\usepackage{bm}
\usepackage{mathtools}

\usepackage{subfigure}
\usepackage{mathrsfs}
\usepackage{bibspacing}
\graphicspath{ {./fig/} }
\usepackage{enumitem}

\def\BibTeX{{\rm B\kern-.05em{\sc i\kern-.025em b}\kern-.08em
    T\kern-.1667em\lower.7ex\hbox{E}\kern-.125emX}}
\begin{document}

\title{Long-Term Behavioral Evaluation for Trusted Collaborator Selection via Bidirectional Mamba} 
 
\author{\IEEEauthorblockN{Botao Zhu and Xianbin Wang}
\IEEEauthorblockA{Dept. of Electrical and Computer Engineering, Western University,
London, Ontario N6A 3K7 CANADA \\
}}




\maketitle

\begin{abstract}
Effective selection of trustworthy collaborators is crucial to ensuring the successful completion of collaborative tasks, which requires accurate assessments of both long-term device behavior and short-term collaborative dynamics. Consistent device behavior patterns, which are learned from historical collaborations, can be used to predict their reliability in future collaborations. However, accurately assessing device behavior based on historical collaborations remains challenging. First, behavior assessment from limited historical collaborations captures only instantaneous past behavior, failing to represent the devices’ true behavior. Second, due to the temporal dependencies of device behavior, a unidirectional evaluation that relies only on earlier collaborations loses the opportunity to learn from subsequent collaborations. Addressing these challenges requires evaluating device behavior based on long-term collaborations while considering both forward and backward temporal dependencies. To this end, this work proposes a bidirectional Mamba-enabled model (BM) for long-term behavioral evaluation. For each short time slot, a graph is constructed among devices based on historical collaborations, and device behavioral features within the slot are then aggregated accordingly. Subsequently, a bidirectional Mamba model integrates these short-term representations across all time intervals, producing a stable and reliable long-term behavior evaluation for each device. Experimental results demonstrate that BM achieves higher evaluation accuracy than baseline methods, thereby enabling the selection of collaborators that maximize the value of task completion.
\end{abstract}


\section{Introduction}

With the increasing complexity of modern applications and interconnected systems, individual devices often face challenges in handling computation-intensive tasks due to limited processing power and energy constraints. To overcome this limitation, distributed resource scheduling has emerged as an effective approach, allowing tasks to be offloaded to networked devices with greater computational capabilities~\cite{10546264}. For example, in vehicular networks, autonomous vehicles can delegate computation-intensive tasks to nearby roadside units to reduce the burden on on-board processors~\cite{9978606}. Similarly, in industrial IoT scenarios, sensor nodes frequently offload data processing tasks to edge gateways~\cite{9833361}. These examples underscore the critical importance of selecting reliable collaborators to ensure successful task execution.

Trust has become a critical measure for evaluating the reliability of devices in collaborative systems, reflecting a task owner’s confidence in a collaborator’s ability to successfully accomplish a task~\cite{chain_of_trsut}. Accurate evaluation of device trust requires assessing both long-term device behavior and short-term collaborative dynamics. Historical collaborations contain implicit behavioral patterns of devices, reflecting whether their performance tends to be reliable or unreliable. Such patterns enable us to predict their reliability in future collaborative tasks. Various methods have been proposed recently to evaluate the historical behavior of collaborators.
For instance, in~\cite{8769947}, trust was assessed for sensor nodes based on their performance in data collection and communication activities. In~\cite{6519238}, the authors evaluated the trustworthiness of devices in heterogeneous wireless networks by collecting multidimensional interaction information, such as packet forwarding success rate and session interruption rate. In~\cite{10571576}, the authors employed a time-window-triggered mechanism to periodically update node trust, focusing on capturing recent changes in node behavior. However, these approaches either rely on static assessments of device behavior or emphasize only short-term conditions, which makes it difficult to accurately reflect their actual reliability.

A device's behavioral patterns are formed through its performance across multiple short-duration collaborations. Therefore, it is necessary to continuously observe and evaluate its long-term behavior to more accurately reflect its reliability. Conducting such long-term evaluation, however, introduces several challenges that must be addressed. First, due to the complexity of collaborations among devices, an effective mechanism is required to infer each device's behavior within a time slot based on inter-device collaborations. Second, the behavior features of a device across time slots are correlated and display both forward and backward temporal dependencies, requiring a fusion mechanism to capture these dependencies for accurate long-term behavioral assessment. Some works used Long Short‑Term Memory (LSTM)-based models to aggregate features across time~\cite{AlghofailiRassam2022}. While LSTM is effective in capturing short-term temporal dependencies, its sequential computation prevents parallel processing. Consequently, as the sequence length grows, its efficiency degrades and its ability to model long-range dependencies becomes limited. Other studies adopted attention-based architectures, such as Transformer, to enhance long-term sequence modeling~\cite{WangYanLanBertinoPedrycz2024}. However, their quadratic computational complexity with respect to sequence length results in substantial resource demands and latency, making them impractical for long-term behavioral evaluation.

Mamba, introduced in 2023~\cite{gu2023mamba}, is a neural network framework designed to efficiently handle long sequential data. Leveraging the selective state space mechanism (SSM), it adaptively identifies and preserves critical state information, allowing the model to effectively capture both short-term and long-term dependencies in a bidirectional manner. Its linear computational complexity enables the processing of extended sequences with high efficiency, making Mamba a suitable solution for long-term behavioral evaluation in this work.


Based on the aforementioned challenges and the advantages of Mamba, this work proposes a bidirectional Mamba-enabled (BM) long-term behavioral evaluation model. The long time period is first divided into multiple short time slots, and a collaboration graph among devices is constructed for each slot based on historical collaboration records. 
Within each graph, devices' historical reliability is aggregated according to their collaboration relationships, capturing short-term behavioral characteristics. Furthermore, the bidirectional Mamba performs forward and backward scans over the historical reliability across all time slots to produce accurate long-term assessments. The main contributions of this paper are summarized as follows.
\begin{itemize}[leftmargin=*]
    \item We develop a long-term behavioral evaluation framework that fully leverages devices’ historical behavior to yield accurate assessments for collaborator selection.

    \item  We design an effective mechanism to integrate devices’ historical reliability based on their collaboration relationships within each short time slot, thereby accurately obtaining short-term behavior assessments. 

    \item We design a bidirectional Mamba-based cross-time-slot fusion model that efficiently captures the temporal dependencies in device behavior over long periods, overcoming the low-efficiency issues of conventional sequential models when handling long sequences.
\end{itemize}

\section{System Model and Problem Description}

We consider a collaborative system comprising a set of devices $\bm{K} = \{k_1,\dots, k_I\}$. Each device can function either as a task owner, producing computational tasks, or as a collaborator, executing tasks from other devices. A device $k_j \in \bm{K}$ is characterized by the tuple $(k_j^{\text{cpu}}, k_j^{\text{re}}, k_j^{\text{tr}}, k_j^{\text{ge}})$, where $k_j^{\text{cpu}}$ denotes its CPU frequency, $k_j^{\text{re}}$ represents the reception power, $k_j^{\text{tr}}$ denotes the transmission power, and $k_j^{\text{ge}}$ is the coordinate in physical space. To monitor collaborations, a dedicated device collects performance indicators from participating collaborators. All observed collaborations over a period $\mathbb{T}$ are stored in a dataset $\bm{F}$. Each entry $f_{(k_i,k_j)} \in \bm{F}$ corresponds to an event where device $k_j$ assists device $k_i$ in completing a task, including relevant performance indicators such as task transmission results, task computation results, and others. We assume that device $k_i$ generates a task $\theta = (\theta^{\text{size}}, \theta^{\text{des}}, \theta^{\text{trust}})$, where $\theta^{\text{size}}$ denotes the task size, $\theta^{\text{des}}$ represents the processing density (cycles/bit), and $\theta^{\text{trust}}$ indicates the minimum trust threshold for potential collaborators. Due to limited computational resources, device $k_i$ offloads the task $\theta$ to a reliable collaborator for execution. 

\subsection{Trust Model}
 A device’s trustworthiness depends on its past collaborative behavior and available resources. The trustworthiness of device $k_j$ as evaluated by device $k_i$ is defined as:
\vspace{-0.05 in}
\begin{align}
    T_{(k_i,k_j)}  =  T^{\text{beh}}_{(k_i,k_j)}(\bm{F}_{k_j})T^{\text{res}}_{(k_i,k_j)}(\theta),
\end{align}
where $\bm{F}_{k_j} \in \bm{F}$ represents the set of historical collaboration records associated with device $k_j$, $T^{\text{beh}}_{(k_i,k_j)} \in [0,1]$ denotes the historical reliability, and $T^{\text{res}}_{(k_i,k_j)} \in [0,1]$ is the task-specific resource trustworthiness.

\subsection{Task Transmission and Task Computation Models}
Once the trust evaluation is completed, devices whose trust scores exceed the minimum required threshold are considered as potential collaborators. Suppose that device $k_j$ is selected as the final collaborator. Then, the achievable transmission rate between the task owner $k_i$ and device $k_j$ is computed as:
\begin{align}
   \gamma_{(k_i,k_j)} = W^{\text{band}}\log_2\left(1 + \frac{k_i^{\text{tr}}g_{(k_i,k_j)}}{N_0}\right), 
\end{align}
where $W^{\text{band}}$ represents the channel bandwidth, $k_i^{\text{tr}}$ is the transmission power of $k_i$, $N_0$ is the noise power, $g_{(k_i,k_j)}$ denotes the channel gain between devices $k_i$ and $k_j$. A simple channel model is adopted: $g_{(k_i,k_j)} = |k_i^{\text{ge}} - k_j^{\text{ge}}|^{-\alpha_0}$, where $|k_i^{\text{ge}} - k_j^{\text{ge}}|$ is the distance between devices, and $\alpha_0 = 4$ is the path loss factor. Based on the transmission rate, the task transmission time and the corresponding energy consumption are approximated as follows:~\cite{6518637}:
\begin{align}
   t^{\text{tr}}_{(k_i,k_j)} &=  \theta^{\text{size}}/\gamma_{(k_i,k_j)}, \\
   E^{\text{tr}}_{(k_i,k_j)} &= t^{\text{tr}} (k_i^{\text{tr}} + k_j^{\text{re}}), 
\end{align}
where $t^{\text{tr}}_{(k_i,k_j)}$ is the task transmission time, and $E^{\text{tr}}_{(k_i,k_j)}$ represents the sum of the transmission energy consumed by the task owner $k_i$ and the reception energy consumed by device $k_j$. Subsequently, device $k_j$ executes the task $\theta$ using its local computational resources. The computation time and energy are expressed as~\cite{11096939}:
\begin{align}
     \label{computation_time}
    t^{\text{com}}_{k_j} &= \theta^{\text{size}} \theta^{\text{des}} / k_j^{\text{cpu}}, \\ 
    \label{computation_ener}
    E^{\text{com}}_{k_j} &= \epsilon (k_j^{\text{cpu}})^2 \theta^{\text{size}} \theta^{\text{des}}, 
\end{align}
where $\epsilon (k_j^{\text{cpu}})^2$ is the coefficient denoting the consumed energy per CPU cycle, and $\epsilon$ is set to $10^{-11}$ according to the measurements in \cite{6195685}. 
Therefore, the total task time for device $k_j$ to execute the task $\theta$ is the sum of the task transmission time and the task computation time, expressed as $t_{k_j}^{\text{tot}} = t^{\text{tr}}_{(k_i,k_j)} + t^{\text{com}}_{k_j}$. Likewise, the total energy consumption in this process is calculated as $E_{k_j}^{\text{tot}} = E^{\text{tr}}_{(k_i,k_j)} + E^{\text{com}}_{k_j}$.

\subsection{Value as a Metric}
To measure the task owner’s satisfaction with device $k_j$ when executing the task $\theta$, we introduce the value of task completion (VoC) as a metric to quantify the execution outcome, which is defined as:
\begin{align}
    V_{k_j} = \xi_1 V^{\text{time}}_{k_j} + \xi_2 V^{\text{ener}}_{k_j},
\end{align}
where $\xi_1$ and $\xi_2$ are the weight parameters, $0 \leq \xi_1, \xi_2 \leq 1$, $\xi_1 + \xi_2 = 1$. The term $V^{\text{time}}_{k_j}$ quantifies the satisfaction from the task completion time perspective. Following the Kano satisfaction model, it is given by~\cite{11096939}:
\begin{align}
    V^{\text{time}}_{k_j} = \begin{cases}
    1, & \text{if} \  t^{\text{com}}_{k_i} \ge t^{\text{tot}}_{k_j}; \\
    e^{-|{(t^{\text{tot}}_{k_j} - t^{\text{com}}_{k_i})}/{t^{\text{com}}_{k_i}}|}, & \text{if} \ t^{\text{com}}_{k_i} < t^{\text{tot}}_{k_j}, \\
   \end{cases}
\end{align}
where $t^{\text{com}}_{k_i}$ denotes the task time if the task $\theta$ is executed locally by the task owner $k_i$, which is obtained from Eq.~(\ref{computation_time}). If $t^{\text{tot}}_{k_j}$ is greater than $t^{\text{com}}_{k_i}$, then $V^{\text{time}}_{k_j}$ is less then 1 and decreases as $t^{\text{tot}}_{k_j}$ increases. Similarly, $V^{\text{ener}}_{k_j}$ measures the satisfaction from the energy consumption perspective, expressed as:
\begin{align}
    V^{\text{ener}}_{k_j} = \begin{cases}
    1, & \text{if} \  E^{\text{com}}_{k_i} \ge E^{\text{tot}}_{k_j}; \\
    e^{-|{(E^{\text{tot}}_{k_j} - E^{\text{com}}_{k_i})}/{E^{\text{com}}_{k_i}}|}, & \text{if} \ E^{\text{com}}_{k_i} < E^{\text{tot}}_{k_j}, \\
   \end{cases}
\end{align}
where $E^{\text{com}}_{k_i}$ is the energy consumption if the task $\theta$ is executed by the task owner $k_i$, which is calculated using Eq.~(\ref{computation_ener}).

\subsection{Problem Formulation}
It can be observed that the VoC depends on which collaborator is selected. Accordingly, this study aims to identify a trusted collaborator to execute the task $\theta$ for the task owner $k_i$ that maximizes the VoC:
\begin{align}
    &\max_{\bm{K}} {V_{k_j}}, \\
    \mathrm{s.t.} \quad 
    &T_{(k_i,k_j)} \geq \theta^{\text{trust}}, \forall k_j \in \bm{K}, k_j \ne k_i \label{1} \\
    & \bm{F}_{k_j} \in \bm{F} \label{2}.
\end{align}
Constraint (\ref{1}) states that the trustworthiness of the selected collaborator should meet the minimum trust threshold $\theta^{\text{trust}}$. Constraint (\ref{2}) specifies that the historical evaluation should be entirely based on the collected historical collaboration data. As indicated by the formulated problem, accurately assessing device trustworthiness from long-term historical collaboration data is essential for achieving reliable collaborator selection.

\section{Bidirectional Mamba-Enabled Long Term Behavioral Evaluation}

\begin{figure}[t!]
\centering
\includegraphics[scale=1.1]{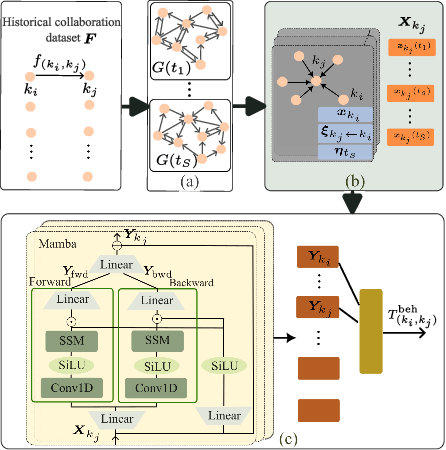}
\caption{The proposed BM model. (a) Constructing a sequence of historical collaboration graphs. (b) Performing fine-grained reliability fusion for devices within each time slot based on historical interactions. (c) Using Mamba to evaluate devices’ long-term historical reliability.}
\label{systemmodel}
\end{figure}

To enable accurate collaborator selection, this study proposes the BM model for long-term behavioral evaluation, as shown in Fig.~\ref{systemmodel}. The model captures a device’s fine-grained historical behavior within each short time slot based on its collaboration relationships. Short-term behavioral features are then fused across the entire time horizon to derive a long-term behavioral evaluation for each device. This section first briefly introduces the Mamba, and then provides a detailed description of the proposed BM model.

\subsection{Mamba}
The Mamba model is a neural architecture recently introduced to capture long-range temporal dependencies efficiently, while keeping computational costs linear, providing a practical alternative to attention-based models like Transformer. It builds on the SSM mechanism, which treats sequences as continuous-time dynamical systems. By leveraging the state-space representation, Mamba incorporates a selective mechanism that adaptively controls how information is retained or discarded at each time step. In this continuous-time formulation, the hidden state $\bm{h}(t)$ and output $\bm{y}(t)$ evolve according to:
\begin{align}
  \label{ssm}
    \bm{h}^{'}(t) &=  A\bm{h}(t) + B\bm{x}(t),\\
    \bm{y}(t) &= C\bm{h}(t),
\end{align}
where $\bm{x}(t)$ denotes the input sequence, and $A$, $B$, $C$ are learnable parameter matrices. For scenarios with discrete inputs, the continuous-time SSM in Eq.~(\ref{ssm}) is discretized using the zero-order hold method~\cite{Ding2025DyGMamba}, yielding the following discrete formulation:
\begin{align}
   \label{ssm2}
    \bm{h}(t) &= \overline{A}\bm{h}(t-1) + \overline{B}\bm{x}(t), \\ 
    \bm{y}(t) &= C \bm{h}(t),
\end{align}
 where $\overline{A} = \exp{(\Delta A)}$, $\overline{B} = (\Delta A)^{-1}(\exp{(\Delta A)}-I)(\Delta B)$, $\Delta$ is a specified sampling timescale for the discretization.

\subsection{Historical Collaboration Graph Sequence Construction}
A device’s historical reliability is evaluated based on the collaborative behavior observed by its collaborating devices. To represent these collaboration relationships, a sequence of historical collaboration graphs is constructed across multiple short time slots. Specifically, the continuous time axis $\mathbb{T}$ is divided into $S$ equal-length time slots, denoted as $\{t_1, \dots,  t_S\}$. Correspondingly, the historical collaboration dataset $\bm{F}$ is partitioned into subsets $\{\bm{F}(t_1), \dots, \bm{F}(t_S)\}$, each representing the collaborations occurring within a specific slot. For each time slot $t_s$, a historical collaboration graph $G(t_s)$ is constructed to model collaboration dependencies among devices based on the corresponding subset $\bm{F}(t_s)$. An edge $e_{(k_i,k_j)}$ from device $k_i$ to device $k_j$ is added if device $k_j$ assists $k_i$ in completing a task during $t_s$. The edge weight quantifies the historical reliability of device $k_j$ as evaluated by device $k_i$ based on their direct collaborations, computed as:
\begin{align}
    T_{(k_i, k_j)}^{\text{col}} =  \frac{1}{N_{(k_i,k_j)}}\sum_{n=1}^{N_{(k_i,k_j)}} \left(\alpha_1 p_n^{\text{\text{tr}}} + \alpha_2 p_n^{\text{com}}\right), 
\end{align} 
where $p^{\text{tr}}_n$ indicates task transmission success ($1$) or failure ($0$) for the $n$-th collaboration in time slot $t_s$, and $p^{\text{com}}_n$ indicates task computation success ($1$) or failure ($0$). $N_{(k_i,k_j)}$ is the number of tasks received by device $k_j$ from device $k_i$.
The parameters $\alpha_1$ and $\alpha_2$ are weights satisfying  $\alpha_1 + \alpha_2 = 1$ and $0 \leq \alpha_1, \alpha_2 \leq 1$. Therefore, we obtain the historical collaboration graph in $t_s$, denoted as $G(t_s) = (\bm{K}(t_s), \bm{E}(t_s), \bm{W}(t_s))$, where $\bm{K}(t_s)$ denotes the set of devices participating in collaborations during time slot $t_s$, $\bm{E}(t_s)$ represents the set of edges capturing pairwise interactions among devices, and $\bm{W}(t_s)$ represents the edge weight set computed based on $\bm{F}(t_s)$. By arranging the graphs from all time slots chronologically, a sequence of collaboration graphs is obtained, denoted as $\{G(t_1), \dots, G(t_S)\}$.

\subsection{Short-Term Behavior Fusion}
In this section, we calculate the historical reliability of each device in each time slot based on the assessments of all devices associated with it.
During time slot $t_s$, when device $k_j$ acts as a collaborator assisting other devices with task execution, the devices it assists collectively determine device $k_j$’s historical reliability in this time slot. In the corresponding graph $G(t_s)$, these assessing devices correspond to the one-hop in-degree neighbors of device $k_j$. To enable devices to perceive each other’s historical reliability, a Graph Neural Network (GNN) is employed to propagate and fuse reliability information across the network graph. For each neighbor $k_i$, the historical reliability towards $k_j$ is encoded as: 
\begin{align}
     \bm{\xi}_{k_j \gets k_i } = W_{k_j \gets k_i} \bm{\psi}_{k_j \gets k_i},\, k_i \in \mathcal{N}_{k_j}, 
\end{align}
where $W_{k_j \gets k_i} \in \mathbb{R}^{d_T \times  d_T}$ is a learnable weight matrix, $\mathcal{N}_{k_j}$ denotes the set of one-hop in-degree neighbors of device $k_j$, and $\bm{\psi}_{k_j \gets k_i} \in \mathbb{R}^{d_T}$ represents the binary-encoded embedding of $T^{\text{col}}_{(k_i,k_j)}$. Following ~\cite{li2025dygmamba}, time slot $t_s$ is encoded as $\bm{\eta}_{t_s} = \sqrt{\frac{1}{d_t}}[\cos(W_1 t_s), \sin(W_1 t_s), \dots, \cos(W_{d_t}t_s), \sin(W_{d_t}t_s)]$,
 where $W_{1}, \dots, W_{d_t}$ are the trainable parameters, $d_{t}$ is the encoding dimension. Then, a learnable linear projection is applied to map both $\bm{\xi}_{k_j \gets k_i}$ and $\bm{\eta}_{t_s}$ to the same dimension $d_{a}$. For simplicity, the same notations are retained to represent the projected embeddings. The message from $k_i$ to $k_j$ is constructed by concatenating these embeddings:
\begin{align}
    \bm{\mu}_{k_j \gets k_i} = \bm{x}_{k_i} \oplus \bm{\xi}_{k_j \gets k_i } \oplus \bm{\eta}_{t_s},
\end{align}
where $\bm{x}_{k_i} \in \mathbb{R}^{d_a}$ is the device embedding generated via node2vec, and $\oplus$ denotes concatenation. Intuitively, $\bm{\mu}_{k_j \gets k_i}$ represents the recommendation from device $k_i$ to device $k_j$ within time slot $t_s$. After collecting messages from all in-degree neighbors, $k_j$ aggregates them using an aggregation function:
\begin{align}
   \bm{x}_{k_j} = \text{AGG}(\bm{\mu}_{k_j \gets k_i}, k_i \in \mathcal{N}_{k_j}). 
\end{align}
To allow reliability to propagate across multiple hops in the graph, $L$ layers of propagation and aggregation are stacked, producing the final embedding of device $k_j$ at time slot $t_s$, denoted as $\bm{x}_{k_j}(t_s)$. By concatenating $k_j$' embeddings from all time slots, a comprehensive representation $\bm{X}_{k_j} = [\bm{x}_{k_j}(t_1); \dots; \bm{x}_{k_j}(t_S)] \in \mathbb{R}^{S \times d_a}$ is constructed, capturing the temporal dynamics of its historical reliability over the time period $\mathbb{T}$. This process is applied to all devices to generate historical reliability embeddings that capture their historical behavior.

\subsection{Bidirectional Mamba–Aided Long-Term Behavior Fusion}

To accurately reflect the historical reliability of devices over an extended time span, the temporal embeddings obtained from the previous stage require effective integration. We leverage Mamba to implement a long-term historical evaluation fusion model that captures both forward and backward temporal dependencies, yielding precise historical assessment results. Each $\text{Mamba}$ layer in our design consists of a forward $\text{Mamba}$ block and a backward $\text{Mamba}$ block. In the forward Mamba block, device $k_j$'s embedding $\bm{X}_{k_j}$ undergoes the following processing pipeline:

\textbf{Input Preprocessing}: The embedding $\bm{X}_{k_j}$ is initially processed through the linear layer and the $\text{1D}$ convolution, followed by the $\text{SiLU}$ activation function to generate the intermediate features $\widetilde{\bm{X}}_{k_j}$, as follows: 
\begin{align}
   \widetilde{\bm{X}}_{k_j}  &= \text{SiLU}(\text{Conv1D}(\text{Linear}(\bm{X}_{k_j}))).
\end{align}
\textbf{Discretization}: The discretized $\overline{A}$ and $\overline{B}$ are obtained by the following steps~\cite{li2025dygmamba}: 
\begin{align}
    B &= \text{Linear}(\widetilde{\bm{X}}_{k_j}), \,
    C = \text{Linear}(\widetilde{\bm{X}}_{k_j}), \\
    \Delta &= \text{Softplus}(\text{Linear}(\widetilde{\bm{X}}_{k_j})),\\
    \overline{A} &= \text{Discrete}( \Delta, A),\,
     \overline{B} = \text{Discrete}(\Delta, A, B),
\end{align}
where $\text{Softplus}(\cdot)$ is a smooth approximation of ReLU function, and $\text{Discrete}(\cdot)$ is the discretization process. 

\textbf{Selective Scan}: The core SSM module $\text{SSM}_{(\overline{A}, \overline{B}, C)}$ applies a recursive scan over $\widetilde{\bm{X}}_{k_j}$ to produce the temporal output $\bm{Y}_{\text{SSM}}$. The final output $\bm{Y}_{\text{fwd}}$ of the forward Mamba block is generated as follows: 
\begin{align}
   \bm{Y}_{\text{SSM}} &= \text{SSM}_{(\overline{A}, \overline{B}, C)}(\widetilde{\bm{X}}_{k_j}), \\
   \bm{Y}_{\text{fwd}} &=  \text{Linear}(\bm{Y}_{\text{SSM}} \odot \text{SiLU}(\text{Linear}(\bm{X}_{k_j}))).
\end{align}  
As indicated in Eq.~(\ref{ssm2}), the SSM follows a recursive formulation, in which the hidden state 
$\bm{h}(t)$ at each step depends on the preceding state 
$\bm{h}(t-1)$ as well as the current input. This recursive mechanism allows the model to integrate both the latest input and the information accumulated from prior steps. We use $\overrightarrow{\text{Mamba}}(\cdot)$ to represent all operations of the forward Mamba block. The backward Mamba block is used to perform a reverse scan over the elements in $\bm{X}_{k_j}$, formulated as:
\begin{align}
    \bm{Y}_{\text{bwd}} =  \overleftarrow{\text{Mamba}}(\bm{X}_{k_j}).
\end{align}
After obtaining the outputs from the forward and backward Mamba blocks, they are fused and incorporated with the original input $\bm{X}_{k_j}$ through a residual connection to generate the final embedding $\bm{Y}_{k_j} \in \mathbb{R}^{S \times d_a}$, formulated as:
 \begin{align}
     \bm{Y}_{k_j} = \text{Linear}(\bm{Y}_{\text{fwd}} + \bm{Y}_{\text{bwd}}) + \bm{X}_{k_j}.
 \end{align}
To capture the most significant features, we apply a max pooling operation to yield the final embedding ${\bm{Y}}_{k_j} = \text{maxpooling}(\bm{Y}_{k_j}) \in \mathbb{R}^{d_a}$.

By aggregating information over the long-term temporal dimension, each device obtains an embedding that captures its historical behavior characteristics. To evaluate the historical reliability of device $k_j$ from the perspective of device $k_i$ over the long time period $\mathbb{T}$, the embeddings $\bm{Y}_{k_i}$ and $\bm{Y}_{k_j}$ are concatenated and fed into a Multi-Layer Perceptron (MLP), which is given by:
\begin{align}
    \bm{T}_{(k_i,k_j)}^{\text{beh}} = \text{MLP}({\bm{Y}}_{k_i} \oplus {\bm{Y}}_{k_j}), \\
    T^{\text{beh}}_{(k_i,k_j)} = \max(\bm{T}^{\text{beh}}_{k_i,k_j}),
\end{align}
where $\bm{T}^{\text{beh}}_{(k_i,k_j)}$ denotes the output vector produced by the MLP, and  $T^{\text{beh}}_{(k_i,k_j)}$ is the maximum value in $\bm{T}_{(k_i,k_j)}$.

The BM model is trained by minimizing the cross-entropy loss between the computed reliability values and the ground-truth values observed from the historical data.
\begin{align}
    \mathcal{L} = \text{cross\_entropy}(\bm{T}^{\text{col}}, \bm{T}^{\text{beh}}),
\end{align}
where $\bm{T}^{\text{beh}}$ is the set of computed reliability values, and $\bm{T}^{\text{col}}$ is the set of ground-truth values.

\subsection{Task-Specific Resource Trust Evaluation}

Task-specific resource evaluation is an essential aspect of device trust, as it reflects a device’s capability to complete the given task $\theta$. For each potential collaborator $k_j$, this evaluation consists of three components: collaborator willingness, communication resource, and computation resource, as follows:
\begin{align}
    T^{\text{res}}_{(k_i,k_j)}  =  T^{\text{will}}_{k_j}T^{\text{tr}}_{k_j}T^{\text{com}}_{k_j},
\end{align}
where $T^{\text{will}}_{k_j}$ is the evaluation result of $k_j$'s willingness, which is given by:
\begin{align}
    T^{\text{will}}_{k_j} = \begin{cases}
    1, & k_j \, \text{is willing to collaborate}; \\
    0, & \text{otherwise}. \\
   \end{cases}
\end{align}
The assessment of a collaborator's communication resources considers several key factors, including channel quality, available bandwidth, achievable data rate, and transmission reliability reflected by latency, jitter, and packet loss. These factors jointly determine whether the collaborator can support stable and timely task transmission. $T^{\text{tr}}_{k_j}$ is the evaluation result of communication resources of $k_j$, which is given by:
\begin{align}
    T^{\text{tr}}_{k_j} = \begin{cases}
    1, & k_j\text{'s} \, \text{communication resources are trusted}; \\
    0, & \text{otherwise}. \\
   \end{cases}
\end{align}
The assessment of a collaborator’s computational resources considers processing capacity, memory and storage availability, and operational stability. These factors collectively indicate whether the collaborator can reliably and efficiently execute assigned tasks. 
$T^{\text{com}}_{k_j}$ is the evaluation result of $k_j$'s computational resources, which is defined as:
\begin{align}
    \hspace{-0.1 in}T^{\text{com}}_{k_j} = \begin{cases}
    1, & k_j\text{'s} \, \text{computational resources are trusted}; \\
    0, & \text{otherwise}. \\
   \end{cases}
\end{align}
Collaborator $k_j$'s resource trust $T^{\text{res}}_{(k_i,k_j)}$ is set to 1 only if all requirements are met; otherwise, it is set to 0. This approach quantifies the reliability of potential collaborators and provides a basis for task assignment. Based on the combination of long-term historical behavior assessment and task-specific resource evaluation, the task owner first identifies the set of devices whose trust values meet the minimum trust threshold $\theta^{\text{trust}}$. From this set, the device that maximizes the VoC is selected as the final collaborator.

\section{Experimental Analysis}

\subsection{Experimental Settings}
To validate the proposed BM model, we implement a wireless system utilizing the $\text{NS-3}$ discrete-event network simulator with Python bindings. $\text{NS-3}$ is selected for its capacity to accurately simulate realistic network behavior and model distributed computing. We deploy 500 devices. Each device is configured with a transmission power of $100$ mW, a reception power of $80$ mW, and a $\text{CPU}$ frequency randomly selected from the set $\{2, 4, 6\}$ $\text{GHz}$. The wireless channel is configured with a bandwidth of $5$ MHz and a noise power of $-80$ dBm. For the trust evaluation component, the weighting coefficients $\alpha_1$ and $\alpha_2$ are set to ${0.6}$ and ${0.4}$, respectively. The parameters $\xi_1$ and $\xi_2$ are both set to 0.5. The system is loaded with ${10,000}$ tasks, executed sequentially. We focused on the face recognition task, characterized by a default input size of $5$ MB and a processing density of $2,339$ cycles/bit \cite{11096939}. Device performance data is systematically recorded throughout the execution of these tasks. Ground-truth values for all devices are generated based on their recorded historical performance, following the established procedure outlined in~\cite{Favour2025Benchmarking}. The initialized embedding dimension is set to ${128}$. The $\text{GNN}$ component utilized ${L=3}$ propagation and aggregation layers. The output dimensions for these three successive layers are configured as ${32}$, ${64}$, and ${32}$, respectively. The Mamba component is constructed with 3 bidirectional $\text{Mamba}$ layers. The dataset is partitioned into an ${80\%}$ training subset and a ${20\%}$ testing subset. Five-fold cross-validation is performed on the training data, and the training process incorporates early stopping to prevent overfitting. The model is trained on the Lambda Vector workstation. Hyperparameters are tuned by sampling from the following ranges: the learning rate $\{10^{-1}, 10^{-2}, 10^{-3}, 10^{-4}\}$, the $L_2$ regularization coefficient $\{10^{-5}, 10^{-4}\}$, and the dropout rate $\{0, 0.1, 0.3, 0.5, 0.8\}$. Unless otherwise specified, all presented experimental results correspond to the best-performing settings: a learning rate of ${10^{-2}}$, an $L_2$ regularization coefficient of ${10^{-5}}$, and a dropout rate of ${0}$.

\subsection{Comparison of Evaluation Accuracy}

\begin{figure}[!t]
      \centering
      \subfigure[]{\includegraphics[scale=0.435]{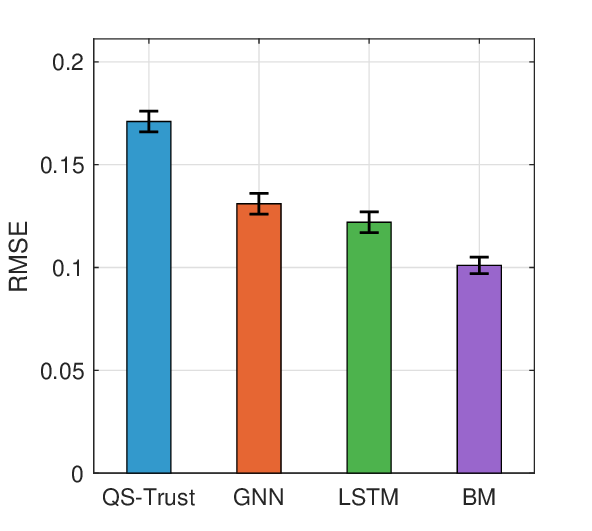}}
      \hspace{-0.01 in}\subfigure[]{\includegraphics[scale=0.435]{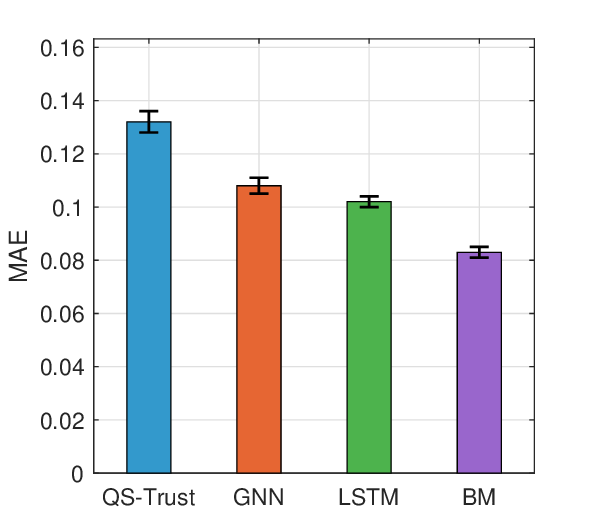}}
      \caption{Comparison of evaluation accuracy. The proposed BM model achieves the lowest RMSE and MAE values.}
     \label{accuracy}
\end{figure}

We first evaluate the accuracy of the proposed BM model in historical reliability assessment. Two widely used metrics–Root Mean Square Error (RMSE) and Mean Absolute Error (MAE)–are employed. Lower values of RMSE and MAE indicate higher accuracy. The comparison results with baseline methods are presented in Fig.~\ref{accuracy}. The results are the average over 10 runs. 
The proposed BM model achieves the lowest RMSE and MAE values, demonstrating high accuracy. This improvement arises from its ability to assess devices' long-term historical reliability based on past collaborations. The performance of LSTM~\cite{AlghofailiRassam2022} surpasses that of GNN~\cite{9488814} and QS-Trust~\cite{Najib2024QSTrust}, yet remains inferior to BM, indicating its limited capability in capturing long-term temporal dependencies. GNN only performs spatial fusion and fails to model long-term information, while QS-Trust relies on a rule-based evaluation scheme without considering temporal dynamics, resulting in the poorest performance.

\begin{figure}[!t]
      \centering
      \subfigure[]{\includegraphics[scale=0.435]{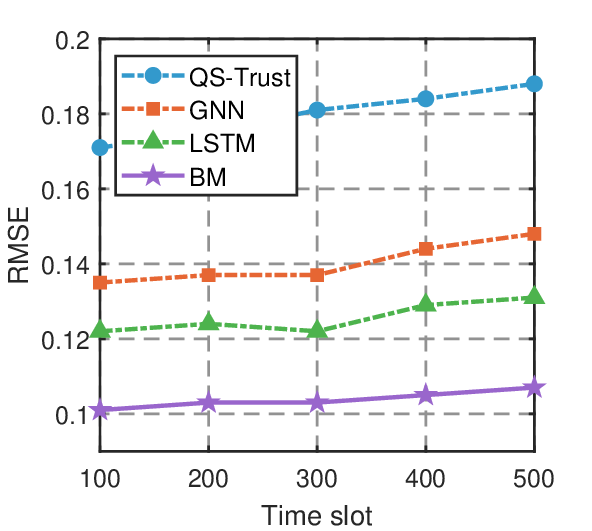}}
      \hspace{-0.01 in}\subfigure[]{\includegraphics[scale=0.435]{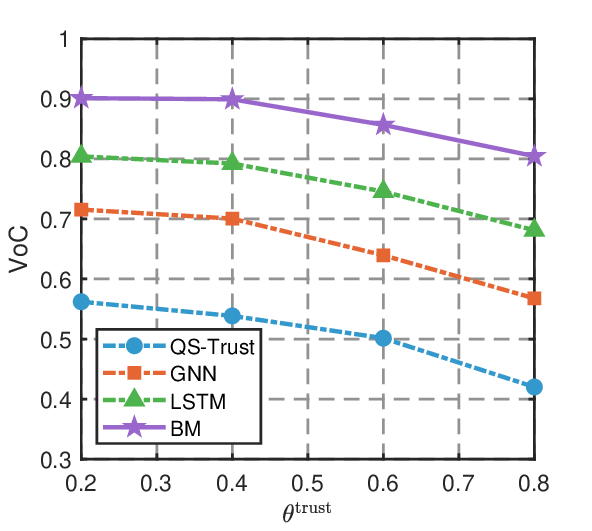}}
      \caption{Both the RMSE comparison and the VoC comparison are conducted over a longer time horizon. (a) The proposed BM model exhibits the smallest fluctuation, reflecting superior long-term performance stability. (b) BM consistently achieves the highest VoC.}
     \label{task}
\end{figure}

\subsection{Long-Term Performance Comparison}
To further evaluate the performance of the proposed model over long time scales, we assess its performance across 500 time slots. As shown in Fig.~\ref{task}~(a), BM exhibits the smallest fluctuation throughout the entire temporal sequence, indicating stable performance in long-term evaluation. Meanwhile, its RMSE values consistently remain lower than those of the comparison algorithms, demonstrating that the proposed BM model maintains high accuracy in long-term inference. In Fig.~\ref{task}~(b), as the minimum trust threshold $\theta^{\text{{trust}}}$ increases, the number of devices that satisfy $\theta^{\text{{trust}}}$ gradually decreases, resulting in a downward trend in the VoC values for all algorithms. However, BM consistently achieves the highest VoC across all threshold levels, indicating its superior capability in accurately identifying trustworthy collaborators.


\vspace{0.1 in}
\section{Conclusion}

This work has investigated the problem of accurately evaluating the historical behavior of devices over long-term periods to support collaborator selection. To address this problem, the BM model is proposed. It first captures the short-term historical reliability of devices based on past collaboration relationships. Then, the bidirectional Mamba is employed to fuse historical reliability across all time slots to produce stable long-term reliability evaluation for all devices. Experimental results demonstrate that BM achieves higher accuracy and stability in long-term behavioral evaluation, thereby supporting the selection of collaborators that maximize the VoC. By enabling reliable and adaptive collaborator selection in dynamic and complex environments, this model paves the way for more robust, intelligent, and trustworthy collaborative systems.


\footnotesize

\end{document}